\documentclass[11pt]{article}
\usepackage{amssymb}
\usepackage[labelfont=bf]{caption}
\usepackage{adjustbox}
\usepackage{array}
\usepackage{makecell}
\newcolumntype{R}[1]{>{\centering\arraybackslash}m{#1}}
\usepackage[english]{babel}
\usepackage{hyphenat}
\usepackage{pgfkeys}
\usepackage{float}
\usepackage{tcolorbox}
\usepackage{enumitem}

\usepackage[preprint]{acl}

\usepackage{times}
\usepackage{latexsym}
\usepackage{amsmath}
\usepackage{pgfplots}
\usepackage{tikz}
\usetikzlibrary{positioning, arrows.meta, shapes.geometric, calc, backgrounds} 
\pgfdeclarelayer{background} 
\pgfsetlayers{background,main} 

\usepackage[T1]{fontenc}

\usepackage[utf8]{inputenc}

\usepackage{microtype}

\usepackage{inconsolata}

\usepackage{graphicx}

\title{Diffutron: A Masked Diffusion Language Model for Turkish Language}

\author{Şuayp Talha Kocabay \\
  Hugging Face: \href{https://huggingface.co/suayptalha}{huggingface.co/suayptalha} \\
  \texttt{kocabaysuayptalha08@gmail.com} \\\And
  Talha Rüzgar Akkuş \\
  Hugging Face: \href{https://huggingface.co/Q-bert}{huggingface.co/Q-bert}\\
  \texttt{talharuzgarakkus@gmail.com} \\}

\begin{document}
\maketitle
\begin{abstract}
Masked Diffusion Language Models (MDLMs) have emerged as a compelling non-autoregressive alternative to standard large language models; however, their application to morphologically rich languages remains limited. In this paper, we introduce \textit{Diffutron}, a masked diffusion language model specifically designed for Turkish. Our approach leverages a resource-efficient training pipeline, starting with LoRA-based continual pre-training of a multilingual encoder on a large-scale corpus. To enable generative capabilities, we employ a progressive instruction-tuning strategy, sequentially adapting the model on general and task-specific instruction sets. Experimental results across comprehensive benchmarks demonstrate that, despite its compact size, our model achieves competitive performance compared to existing multi-billion-parameter baselines. These findings validate the effectiveness of masked diffusion modeling combined with multi-stage tuning for non-autoregressive text generation in Turkish.

\end{abstract}

\section{Introduction}
Autoregressive (AR) Transformers currently dominate the landscape of Large Language Models (LLMs), driving the success of influential models like GPT and Llama \cite{vaswani2017attention, brown2020language, touvron2023llamaopenefficientfoundation}. However, their fundamental nature of generating text one token at a time imposes limitations on generation speed and restricts the context they can consider. Recently, Masked Diffusion Language Models (MDLMs) have emerged as a compelling non-autoregressive alternative, generating text by iteratively refining it and allowing for simultaneous consideration of the entire sentence context \cite{sahoo2024simpleeffectivemaskeddiffusion}. \\

Despite these advantages, research on diffusion models has largely focused on English, leaving their effectiveness on morphologically rich, agglutinative languages like Turkish not well understood. Applying these new architectures to such languages presents unique challenges concerning training stability and specific data requirements. To address this critical gap, we introduce \textit{Diffutron}, a lightweight and parameter-efficient masked diffusion language model specifically tailored for Turkish. Our work provides one of the first detailed applications of this architecture to an agglutinative language. We employ a multi-stage training pipeline that begins with Low-Rank Adaptation (LoRA)-based continual pre-training of the \textit{jhu-clsp/mmBERT-base} multilingual encoder on a large-scale corpus \cite{marone2025mmbertmodernmultilingualencoder}. Importantly, LoRA is used to preserve the core linguistic knowledge of the base model while adapting it to Turkish. \\

Crucially, to unlock high-quality generative capabilities, we propose a progressive instruction-tuning strategy, sequentially adapting the model on general and task-specific instruction sets for greater coherence and helpfulness. We used the \textit{dllm} library for instruction tuning \cite{dllm}. \\
Fine-tuning procedure consisted of two main stages. First, we employed the \textit{metunlp/LlamaTurk-Instruction-Set} dataset \cite{toraman2024llamaturkadaptingopensourcegenerative} to teach the model the fundamentals of instruction-following. This initial stage enabled the model to grasp basic patterns of responding to instructions. In the second stage, we trained the model using the \textit{turkish-nlp-suite/InstrucTurca} dataset \cite{instructurca}, which provides more complex instruction examples, thereby enhancing the model's ability to handle intricate commands. Our experimental findings confirm that combining masked diffusion modeling with this multi-stage tuning approach is effective. Evaluations on comprehensive benchmarks, including a representative subset of the CETVEL suite \cite{er2025cetvelunifiedbenchmarkevaluating}, demonstrate that our model achieves competitive performance compared to existing standard autoregressive models. Remarkably, \textit{Diffutron} delivers these results with only 307 million parameters, proving to be significantly more resource-efficient than roughly $7\times$ larger autoregressive baselines (e.g., 2B parameters). These findings validate the effectiveness of masked diffusion modeling combined with multi-stage tuning for non-autoregressive text generation in Turkish, offering a viable path for high-performance generation under constrained computational budgets.

\section{Related Works}

\subsection{Evolution of Diffusion Models in Text Generation} While autoregressive (AR) models dominate text generation, their sequential nature creates bottlenecks in planning and inference. Early diffusion approaches in NLP, such as Diffusion-LM~\cite{li2022diffusion}, embed discrete text into continuous latent spaces and rely on a rounding step to recover tokens, introducing challenges in mapping continuous states back to discrete text. Consequently, the field has shifted toward Discrete Masked Diffusion Models \cite{austin2021structured}, which operate directly on token states via transition matrices, conceptually aligning with the Masked Language Modeling (MLM) objective. Recent scalable implementations like LLaDA \cite{nie2025llada}, Dream 7B \cite{ye2025dream}, and Mercury \cite{grover2025mercury} have demonstrated that MDLMs can rival autoregressive baselines in quality while enabling parallel generation. Our work extends this discrete lineage to the unexplored domain of morphologically rich languages.

\subsection{Instruction Tuning and Continual Adaptation} Instruction tuning aligns language models with human intent, often utilizing synthetic datasets like Alpaca \cite{taori2023alpaca} to mitigate data scarcity. However, adapting multilingual foundations to specific languages poses a risk of catastrophic forgetting, where the model loses general knowledge while optimizing for a new target distribution \cite{li2025rethinking}. Full-parameter fine-tuning often disrupts pre-trained feature spaces. To address this stability-plasticity dilemma, we employ Low-Rank Adaptation (LoRA) \cite{hu2021lora}. In our framework, LoRA serves as a regularization mechanism during Continual Pre-training (CPT), ensuring the model adapts to Turkish linguistic structures while preserving the robust cross-lingual representations of the base encoder \cite{ren2024analyzingreducingcatastrophicforgetting}.

\subsection{Generative Landscape of Turkish NLP} Turkish NLP has evolved from discriminative encoder-only models like BERTurk \cite{schweter2020berturk} to generative architectures. Recent autoregressive models, including Kanarya \cite{kanarya2b} and Kumru \cite{kumru2b}, alongside TURNA \cite{uludogan2024turna}, have established strong baselines for sequential generation. Despite these advancements, the ecosystem remains exclusively dominated by autoregressive paradigms. The potential of non-autoregressive modeling, particularly Masked Diffusion, remains largely unexplored for morphologically rich, agglutinative languages. This study bridges this gap, leveraging benchmarks like Cetvel \cite{er2025cetvelunifiedbenchmarkevaluating} to evaluate the first Turkish Masked Diffusion Language Model.

\section{Preliminaries}
\label{sec:preliminaries}

In this section, we outline the formulation of Masked Diffusion Language Models (MDLMs) utilized in Diffutron. Unlike autoregressive models that generate tokens sequentially ($x_1, x_2, \dots$), MDLMs treat text generation as a discrete diffusion process, generating all tokens in parallel through iterative refinement.

\subsection{Forward Process: Corruption}
The forward process is a Markov chain that gradually corrupts a clean data sample $x_0$ (the original text) into pure noise $x_T$ over $T$ timesteps. In the context of MDLMs, "noise" is represented by a special absorbing state, the \texttt{<mask>} token.

Let $x_t$ denote the sequence of tokens at timestep $t$. The transition from $x_{t-1}$ to $x_t$ is defined by a transition matrix $Q_t$, where each token either remains unchanged or is replaced by \texttt{<mask>} with a probability $\beta_t$:

\begin{equation}
q(x_t | x_{t-1}) = \mathcal{C}(x_t | x_{t-1}, \beta_t)
\end{equation}

Mathematically, for a single token, the transition probability is:
\begin{equation}
q(x_t^i | x_{t-1}^i) = \begin{cases} 
1 - \beta_t & \text{if } x_t^i = x_{t-1}^i \\
\beta_t & \text{if } x_t^i = \texttt{<mask>} \\
0 & \text{otherwise}
\end{cases}
\end{equation}
By the final timestep $T$, the sequence $x_T$ effectively becomes a sequence entirely composed of \texttt{<mask>} tokens.

\subsection{Reverse Process: Denoising}
The generative capability of \textit{Diffutron} comes from learning the reverse process $p_\theta(x_{t-1} | x_t)$, which attempts to denoise the sequence by predicting the original tokens for the masked positions. 

Since the exact reverse posterior $q(x_{t-1} | x_t, x_0)$ is intractable without knowing $x_0$, we approximate it using a neural network trained to predict $x_0$ directly from the noisy state $x_t$. The generative process starts from a fully masked sequence $x_T$ and iteratively samples $x_{t-1}$ using the predicted probabilities:
\vspace{-0.1cm}
\begin{equation}
p_\theta(x_{t-1} | x_t) = \sum_{\tilde{x}_0} q(x_{t-1} | x_t, \tilde{x}_0) p_\theta(\tilde{x}_0 | x_t)
\end{equation}
\vspace{-0.1cm}

\begin{figure}[ht!]
    \centering
    \resizebox{\columnwidth}{!}{%
    \begin{tikzpicture}[
        node distance=0.5cm,
        token/.style={draw=gray!60, fill=white, rounded corners=1pt, minimum height=0.45cm, minimum width=1.1cm, font=\scriptsize\sffamily, align=center, inner sep=2pt},
        mask/.style={token, draw=red!60, fill=red!5, text=red!70, font=\bfseries\scriptsize\sffamily, dashed},
        arrow/.style={-Latex, thick, color=blue!50},
        context_line/.style={-Latex, thin, color=gray!60, bend left=45, dashed},
        label_text/.style={font=\tiny\bfseries, color=gray!80, anchor=east} 
    ]

    \draw[->, thick, gray!40] (-1.5, 0.2) -- (-1.5, -2.5);
    \node[font=\tiny\sffamily\bfseries, color=gray, rotate=90, anchor=south] at (-1.5, -1.15) {Reverse Process};

    \node[label_text] at (-0.9, 0) {$x_T$}; 
    
    \node[mask] (m1) at (0, 0) {<mask>};
    \node[mask] (m2) at (1.25, 0) {<mask>};
    \node[mask] (m3) at (2.5, 0) {<mask>};
    \node[mask] (m4) at (3.75, 0) {<mask>};

    \draw[arrow] (1.875, -0.3) -- (1.875, -0.8);

    \node[label_text] at (-0.9, -1.1) {$x_t$};
    
    \node[token] (p1) at (0, -1.1) {Bugün};
    \node[mask] (p2) at (1.25, -1.1) {<mask>};
    \node[token] (p3) at (2.5, -1.1) {çok};
    \node[mask] (p4) at (3.75, -1.1) {<mask>};

    \draw[context_line, bend left=50] (p1.north) to (p2.north);
    \draw[context_line, bend right=50] (p3.north) to (p2.north);
    \draw[context_line, bend left=50] (p3.north) to (p4.north);

    \draw[arrow] (1.875, -1.4) -- (1.875, -1.9);

    \node[label_text] at (-0.9, -2.2) {$x_0$};
    
    \node[token, fill=green!5, draw=green!40] (c1) at (0, -2.2) {Bugün};
    \node[token, fill=green!5, draw=green!40] (c2) at (1.25, -2.2) {hava};
    \node[token, fill=green!5, draw=green!40] (c3) at (2.5, -2.2) {çok};
    \node[token, fill=green!5, draw=green!40] (c4) at (3.75, -2.2) {güzel};
    
    \begin{pgfonlayer}{background}
        \draw[dashed, gray!20, rounded corners] (-2.0, 0.5) rectangle (4.5, -2.7);
    \end{pgfonlayer}

    \end{tikzpicture}%
    }
    \vspace{-0.2cm}
    \caption{Overview of the reverse diffusion step. The model predicts masked tokens (e.g., $x_t \to x_0$) by attending to visible context bi-directionally (dashed lines).}
    \label{fig:diffusion_process}
\end{figure}
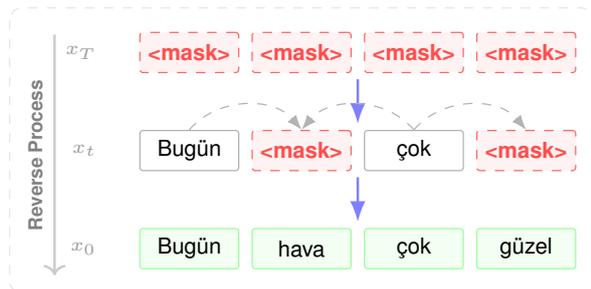

This allows the model to refine the entire sentence globally rather than locally. Figure \ref{fig:diffusion_process} illustrates this reverse generation process specifically for a Turkish sentence.

\section{Continual Pre-training}
\label{sec:continual_pretraining}

To adapt the multilingual representations of the base model to the specific linguistic nuances of Turkish, we conduct a Continual Pre-training (CPT) stage. This phase serves to align the encoder's latent space with the target language distribution while strictly preserving the semantic reasoning capabilities acquired during the initial pre-training. We utilize the \textit{jhu-clsp/mmBERT-base} model as our backbone \cite{marone2025mmbertmodernmultilingualencoder}.

\subsection{Data Curation and Processing}
To align the model's representations with Turkish linguistic dynamics, we curated a composite dataset derived from three primary open-source collections: \textit{Havadis}, \textit{Temiz-OSCAR}, and \textit{Turkish Wikipedia} \cite{10.1007/978-3-031-70563-2_16, wikidump}. Our curation strategy prioritized a balance between encyclopedic knowledge and general web usage while adhering to the context window constraints of the architecture.

We sourced our data from the following repositories:
\begin{itemize}[leftmargin=*, align=left]
    \item \textbf{Havadis:} A comprehensive dataset of Turkish news articles.
    \item \textbf{Temiz-OSCAR:} A filtered and cleaned version of the Common Crawl-based OSCAR corpus \cite{Ortiz_Su_rez_2020}.
    \item \textbf{Turkish Wikipedia:} The standard encyclopedic subset from the Wikimedia Foundation.
\end{itemize}

\paragraph{Preprocessing and Sampling.}
To ensure compatibility with the base model's architecture, we applied a strict length constraint across all data sources, filtering out sequences exceeding a maximum token length of 512. 

The dataset construction proceeded in two phases. First, we processed the Turkish Wikipedia subset with the aforementioned length filter, yielding approximately 406,000 high-quality encyclopedic sequences. Second, we merged the \textit{Havadis} and \textit{Temiz-OSCAR} datasets to form a diverse pool of web and news content. After filtering for length, this merged corpus was shuffled to ensure distributional uniformity. From this pool, we sampled 1.6 million sequences.

The final training corpus consists of the combination of these two subsets, resulting in a total of approximately 2 million sequences (~1.6M web/news and ~406k encyclopedic). We tokenized the final dataset using the base model's tokenizer to maintain alignment with the pre-trained embedding space.

\subsection{Efficient Adaptation via LoRA}
To adapt multilingual representations to Turkish without catastrophic forgetting, we employ Low-Rank Adaptation (LoRA) \cite{hu2021lora}. Unlike standard implementations that limit adaptation to query/value projections, we target all linear modules (Attention Q, K, V, O and MLP Input, Output) to capture the agglutinative complexity of Turkish.

We configured the rank $r=256$ and alpha $\alpha=256$ with a dropout of $0.1$. This results in 14.94\% trainable parameters, a capacity we deem necessary to model Turkish morphological nuances while preserving the frozen backbone's cross-lingual reasoning.

\subsection{Training Configuration and Dynamics}
We trained the model using the Masked Language Modeling (MLM) objective with the memory-efficient \texttt{Paged AdamW 8-bit} optimizer. A cosine learning rate scheduler (peak $5e^{-5}$) was employed. To balance memory usage and training stability, we set the per-device batch size to 64 with 2 gradient accumulation steps, resulting in an effective batch size of 128. Full hyperparameters are listed in Table \ref{tab:training_params}.

\begin{table}[h]
\centering
\small
\caption{Hyperparameters and training statistics for the Continual Pre-training stage.}
\label{tab:training_params}
\setlength{\tabcolsep}{5pt} 
\begin{tabular}{p{4cm} p{3cm}}
\hline
\textbf{Hyperparameter} & \textbf{Configuration} \\
\hline
Base Architecture & mmBERT-base \\
Adapter Rank ($r$) / Alpha ($\alpha$) & 256 / 256 \\
Target Modules & \makecell[l]{Attn ($W_q, W_k, W_v, W_o$) \\ MLP ($W_{up}, W_{down}$)} \\
Trainable Parameters & $\approx$ 14.94\% \\
\hline
Maximum sequence length & 512 \\
Optimizer & Paged AdamW (8-bit) \\
Peak Learning Rate & $5e^{-5}$ \\
Per Device Batch Size & 64  \\
Gradient Accumulation Steps & 2  \\
Effective Batch Size & 128 \\
Precision & Mixed (FP16/BF16) \\
Total Steps & 62,680 (4 Epochs) \\
\hline
\end{tabular}
\end{table}
\vspace{-0.1cm}

Training was completed in approximately 5.9 hours on a single NVIDIA B200 GPU.


\begin{figure}[ht]
    \centering
    \includegraphics[width=0.9\linewidth]{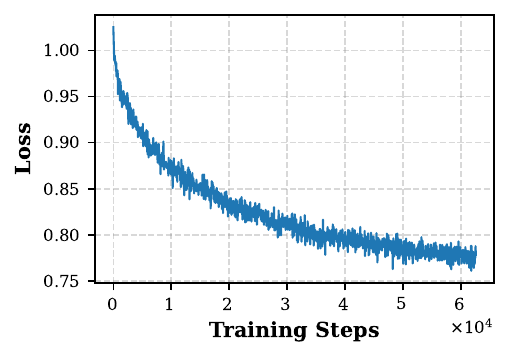}
    \caption{Pretraining loss curve for the model.}
    \label{fig:pretraining-loss}
\end{figure}
\vspace{-0.1cm}

Figure \ref{fig:pretraining-loss} shows a steady decrease in loss. This trajectory confirms that our high-rank adaptation strategy ($r=256$) effectively models the target distribution without training instability.

\section{Instruction Fine-Tuning}
Following continual pre-training, the model underwent a full supervised fine-tuning (SFT) process using a two-stage strategy to progressively enhance instruction-following capabilities. Both stages employed the AdamW optimizer with a learning rate of $1 \times 10^{-4}$, ensuring stable convergence while minimizing overfitting. The two-stage approach was designed to first establish general instruction-following behavior and then specialize in Turkish language instruction tasks.

\subsection{First Stage}
In the first stage, the model was fine-tuned on the \textit{metunlp/LlamaTurk-Instruction-Set} dataset, which consists of diverse instruction-response pairs in Turkish. The goal of this stage was to introduce the model to a broad range of instructions and to improve its general understanding and response coherence. The training configuration, summarized in Table~\ref{tab:first_stage}, used a relatively small batch size to ensure stability during gradient updates over 20 epochs.

\begin{table}[ht]
\centering
\small
\begin{tabular}{ll}
\hline
\textbf{Hyperparameter} & \textbf{Configuration} \\
\hline
Maximum sequence length & 256 \\
Training epochs & 20 \\
Batch size (train/eval) & 16 \\
Optimizer & AdamW (lr=1e-4) \\
Training hardware & A100 40 GB GPU \\
Runtime & 1h 40m \\
\hline
\end{tabular}
\caption{First stage fine-tuning configuration}
\label{tab:first_stage}
\end{table}
\vspace{-0.1cm}

\begin{figure}[H]
    \centering
    \includegraphics[width=0.85\linewidth]{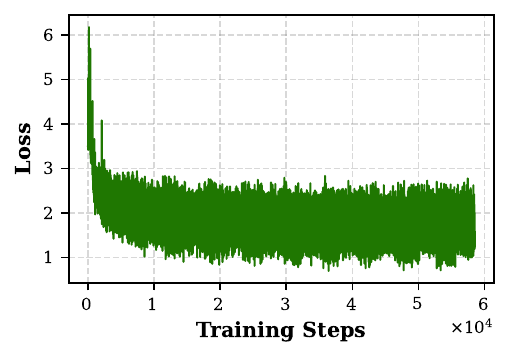}
    \vspace{-0.2cm}
    \caption{Training loss for the first stage of instruction fine-tuning.}
    \label{fig:first-stage-loss}
\end{figure}

The training loss, shown in Figure~\ref{fig:first-stage-loss}, demonstrates a consistent decline over the course of training, indicating effective learning and convergence. Early in the training process, the loss decreased rapidly, reflecting the model’s ability to quickly adapt to the structure and style of instruction-response pairs. Towards the later epochs, the loss plateaued, suggesting that the model had effectively captured the general instruction-following patterns in the dataset. This stage laid the foundation for more specialized fine-tuning in the subsequent stage.


\subsection{Second Stage}
The second stage of fine-tuning focused on the \textit{turkish-nlp-suite/InstrucTurca} dataset, which contains more specialized and nuanced Turkish instruction data. This stage aimed to enhance the model’s performance on more complex or domain-specific instructions, improving its overall utility for Turkish NLP tasks. Training configuration details are provided in Table~\ref{tab:second_stage}. Notably, the batch size was significantly increased and two A100 GPUs were utilized, allowing for faster and more efficient training over 8 epochs.

\begin{table}[h]
\centering
\small
\begin{tabular}{ll}
\hline
\textbf{Hyperparameter} & \textbf{Configuration} \\
\hline
Maximum sequence length & 256 \\
Training epochs & 8 \\
Batch size (train) & 96 \\
Optimizer & AdamW (lr=1e-4) \\
Training hardware & 2xA100 80 GB GPUs \\
Runtime & 9h 53m \\
\hline
\end{tabular}
\caption{Second stage fine-tuning configuration}
\label{tab:second_stage}
\end{table}

\begin{figure}[ht]
    \centering
    \includegraphics[width=0.9\linewidth]{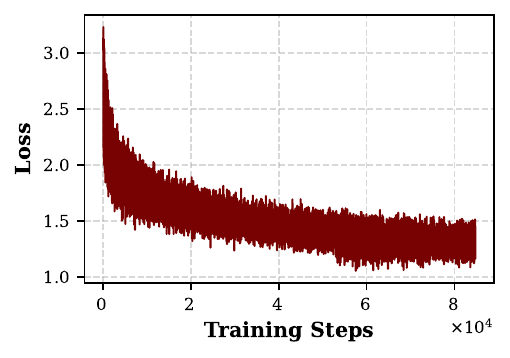}
    \caption{Training loss for the second stage of instruction fine-tuning.}
    \label{fig:second-stage-loss}
\end{figure}

Figure~\ref{fig:second-stage-loss} illustrates the loss progression during this stage. Similar to the first stage, the loss decreased steadily, though the larger batch size and more complex dataset resulted in a smoother and slightly slower convergence curve. The absence of an evaluation strategy was intentional to maximize exposure to the training data and ensure the model adapted fully to the new instruction patterns. By the end of this stage, the model demonstrated improved instruction-following performance, particularly on more intricate or context-sensitive tasks, solidifying its capability as a Turkish instruction-tuned language model.


\section{Evaluation}

In this section, we assess the efficacy of the proposed \textit{Diffutron} through a two-fold evaluation strategy. We first analyze the intrinsic quality of the language modeling improvements gained during continued pre-training, followed by a comparative analysis of downstream performance across a diverse set of Turkish NLP benchmarks.

\subsection{Language Modeling Analysis}
To quantify the improvements gained from the Continued Pre-Training (CPT) phase, we conducted an intrinsic evaluation using perplexity as the primary metric. We utilized the \textit{Bilkent Turkish Writings Dataset} \cite{yilmaz_bilkent_turkish_writings_2025} to assess the models' fluency and adaptation to Turkish linguistic structures. The evaluation was performed with a maximum sequence length of 512 and a masked language modeling (MLM) probability of 0.15.

We compared the perplexity scores of the \textit{jhu-clsp/mmBERT-base} (Pre-CPT) model against the \textit{DiffutronLM-0.3B-Base} (Post-CPT) model. As shown in our analysis, the CPT process resulted in a significant reduction in perplexity:

\begin{itemize}
    \item \textbf{jhu-clsp/mmBERT-base:} 3.42
    \item \textbf{DiffutronLM-0.3B-Base:} 2.75
\end{itemize}

The drop in perplexity from 3.42 to 2.75 indicates that the CPT stage effectively enhanced the model's predictive capabilities and reduced uncertainty, leading to better alignment with the target language distribution.

\subsection{Downstream Task Performance}

When evaluating Turkish large language models (LLMs), benchmark resources remain limited, as many datasets available in Turkish are direct translations of benchmarks originally created for other languages, which do not fully capture the linguistic characteristics of Turkish. After reviewing the available options, we adopted the CETVEL Benchmark Suite due to its structured design and wide applicability. However, running the full CETVEL suite is computationally expensive and time-consuming. Because our budget and computational resources were limited, we selected only the parts that could be feasibly evaluated within our constraints. In this study, we used the following benchmarks: Belebele\_TR for machine reading comprehension \cite{bandarkar2023belebele}, EXAMS\_TR for cross-lingual question answering \cite{hardalov2020exams}, IronyTR for irony detection \cite{ozturk2021ironytr}, News Category Classification \cite{amasyali2004news}, MNLI\_TR for natural language inference \cite{budur2020mnli}, STS\_TR for semantic textual similarity \cite{fikri2021sts}, and XCOPA\_TR for causal commonsense reasoning \cite{ponti2020xcopa}. This subset allowed us to construct a meaningful and computationally manageable evaluation setting.

\begin{table}[ht]
\centering
\caption{Evaluation results on selected CETVEL benchmarks (Sorted by Parameter Count).}
\label{tab:eval-results}
\begin{adjustbox}{max width=\linewidth}
\begin{tabular}{l R{0.6cm} R{0.6cm} R{0.6cm} R{0.6cm} R{0.6cm} R{0.6cm} R{0.6cm} R{0.6cm}}
\hline
\textbf{Benchmark} & 
\makecell{\scriptsize\rotatebox{65}{\textbf{DiffutronLM-0.3B-1st-Stage}}} & 
\makecell{\scriptsize\rotatebox{65}{\textbf{DiffutronLM-0.3B-2nd-Stage}}} & 
\makecell{\scriptsize\rotatebox{65}{boun-tabi-LMG/TURNA}} & 
\makecell{\scriptsize\rotatebox{65}{vngrs-ai/Kumru-2B}} & 
\makecell{\scriptsize\rotatebox{65}{asafaya/kanarya-2b}} & 
\makecell{\scriptsize\rotatebox{65}{Llama-3.2-3B-Instruct}} & 
\makecell{\scriptsize\rotatebox{65}{Trendyol-LLM-7b-base}} & 
\makecell{\scriptsize\rotatebox{65}{CohereForAI/aya-101}} \\
\hline
Belebele\_TR & 22.22 & 27.00 & 22.56 & 29.00 & 28.11 & \textbf{55.78} & 36.22 & 22.89 \\
EXAMS\_TR    & 25.95 & 27.74 & 23.66 & \textbf{30.03} & \textbf{30.03} & 26.21 & 28.50 & 22.90 \\
IronyTR      & 50.67 & \textbf{52.00} & 48.33 & 51.00 & 50.00 & 50.17 & 50.00 & \textbf{52.17} \\
News\_Cat    & 23.20 & 32.40 & 32.80 & 26.40 & 66.80 & 64.00 & \textbf{81.20} & 20.00 \\
MNLI\_TR     & 33.29 & 32.81 & 34.94 & \textbf{36.42} & 33.40 & 34.76 & 35.19 & 27.90 \\
STS\_TR      & 17.77 & \textbf{18.78} & 14.21 & 11.75 & 12.91 & 12.91 & 15.52 & 16.97 \\
XCOPA\_TR    & 53.80 & 52.00 & 55.80 & 54.00 & \textbf{64.20} & 54.60 & 61.00 & 59.60 \\
\hline
\textbf{Average} & 32.41 & \textbf{34.68} & 33.19 & 34.09 & 40.78 & 42.63 & \textbf{43.95} & 31.78 \\
\textbf{Params}  & \textbf{0.3B}  & \textbf{0.3B}  & 1.1B  & 2B    & 2B    & 3B    & 7B    & 13B   \\
\hline
\end{tabular}
\end{adjustbox}
\end{table}

Table \ref{tab:eval-results} presents a comparative analysis of Diffutron against prominent Turkish autoregressive baselines. The most significant finding is the model's efficiency relative to its size. Despite having only 307 million parameters, Diffutron (2nd Stage) achieves an average score of 34.68, surpassing significantly larger models such as Kumru-2B (34.09) and TURNA (33.19). This indicates that the masked diffusion objective is highly effective at compressing linguistic knowledge into a compact latent space. Furthermore, the progression from the 1st to the 2nd stage demonstrates the efficacy of our multi-stage tuning strategy, yielding consistent improvements across semantic tasks like News Classification and STS\_TR.

\section*{Limitations}
Our work is primarily constrained by the current state of the Turkish NLP ecosystem and computational resources. First, the significant lack of modern, native encoder-only foundation models for Turkish necessitated the use of a multilingual backbone, potentially limiting the representational quality compared to a dedicated native architecture. Second, the scarcity of high-quality, native Turkish instruction datasets limits the model's ability to capture complex cultural and linguistic nuances, as existing resources often rely on translations or synthetic data. Additionally, the inherited 256-token context window restricts the model's applicability in long-form generation and summarization tasks. Finally, due to computational constraints, our evaluation was limited to a representative subset of the CETVEL benchmark rather than the full suite.

\section*{Conclusion}

In this study, we presented \textit{Diffutron}, marking a significant step in adapting Masked Diffusion Language Models (MDLMs) to morphologically rich languages like Turkish. By shifting away from the traditional autoregressive framework, we illustrated that alternative generative paradigms can offer robust linguistic capabilities with remarkable parameter efficiency. Our findings confirm that through careful adaptation and tuning, smaller models can effectively bridge the gap with much larger baselines.

We have published our models and dataset on Hugging Face to support further research and development in this field. The models can be accessed through the following links:
\begin{itemize}
    \item \href{https://huggingface.co/diffutron/DiffutronLM-0.3B-Base}{\textit{diffutron/DiffutronLM-0.3B-Base}}
    \item \href{https://huggingface.co/diffutron/DiffutronLM-0.3B-1st-Stage}{\textit{diffutron/DiffutronLM-0.3B-1st-Stage}}
    \item \href{https://huggingface.co/diffutron/DiffutronLM-0.3B-Instruct}{\textit{diffutron/DiffutronLM-0.3B-Instruct}}
\end{itemize}

The pre-training dataset can be accessed here:
\begin{itemize}
    \item \href{https://huggingface.co/datasets/diffutron/DiffutronLM-Pretraining-Corpus}{\textit{diffutron/DiffutronLM-Pretraining-Corpus}}
\end{itemize}

This work challenges the notion that massive scale is the only path to competence in complex languages, highlighting the potential of diffusion-based architectures as viable alternatives. We hope \textit{Diffutron} serves as a catalyst for broader exploration into non-autoregressive modeling, encouraging the community to further investigate diverse architectural approaches for low-resource and agglutinative language processing.

\section*{Acknowledgments}
We gratefully acknowledge the developers of the \textit{dllm} library and KUIS for providing the CETVEL benchmark suite. We also extend our sincere appreciation to Yavuz Alp Sencer Öztürk for his insightful comments and constructive feedback.
\bibliography{custom}

@misc{vaswani2017attention,
      title={Attention Is All You Need}, 
      author={Ashish Vaswani and Noam Shazeer and Niki Parmar and Jakob Uszkoreit and Llion Jones and Aidan N. Gomez and Lukasz Kaiser and Illia Polosukhin},
      year={2023},
      eprint={1706.03762},
      archivePrefix={arXiv},
      primaryClass={cs.CL},
      url={https://arxiv.org/abs/1706.03762}, 
}

@misc{brown2020language,
      title={Language Models are Few-Shot Learners}, 
      author={Tom B. Brown and Benjamin Mann and Nick Ryder and Melanie Subbiah and Jared Kaplan and Prafulla Dhariwal and Arvind Neelakantan and Pranav Shyam and Girish Sastry and Amanda Askell and Sandhini Agarwal and Ariel Herbert-Voss and Gretchen Krueger and Tom Henighan and Rewon Child and Aditya Ramesh and Daniel M. Ziegler and Jeffrey Wu and Clemens Winter and Christopher Hesse and Mark Chen and Eric Sigler and Mateusz Litwin and Scott Gray and Benjamin Chess and Jack Clark and Christopher Berner and Sam McCandlish and Alec Radford and Ilya Sutskever and Dario Amodei},
      year={2020},
      eprint={2005.14165},
      archivePrefix={arXiv},
      primaryClass={cs.CL},
      url={https://arxiv.org/abs/2005.14165}, 
}

@misc{ren2024analyzingreducingcatastrophicforgetting,
      title={Analyzing and Reducing Catastrophic Forgetting in Parameter Efficient Tuning}, 
      author={Weijieying Ren and Xinlong Li and Lei Wang and Tianxiang Zhao and Wei Qin},
      year={2024},
      eprint={2402.18865},
      archivePrefix={arXiv},
      primaryClass={cs.LG},
      url={https://arxiv.org/abs/2402.18865}, 
}

@misc{austin2021structured,
      title={Structured Denoising Diffusion Models in Discrete State-Spaces}, 
      author={Jacob Austin and Daniel D. Johnson and Jonathan Ho and Daniel Tarlow and Rianne van den Berg},
      year={2023},
      eprint={2107.03006},
      archivePrefix={arXiv},
      primaryClass={cs.LG},
      url={https://arxiv.org/abs/2107.03006}, 
}

@misc{li2022diffusion,
      title={Diffusion-LM Improves Controllable Text Generation}, 
      author={Xiang Lisa Li and John Thickstun and Ishaan Gulrajani and Percy Liang and Tatsunori B. Hashimoto},
      year={2022},
      eprint={2205.14217},
      archivePrefix={arXiv},
      primaryClass={cs.CL},
      url={https://arxiv.org/abs/2205.14217}, 
}

@misc{hu2021lora,
      title={LoRA: Low-Rank Adaptation of Large Language Models}, 
      author={Edward J. Hu and Yelong Shen and Phillip Wallis and Zeyuan Allen-Zhu and Yuanzhi Li and Shean Wang and Lu Wang and Weizhu Chen},
      year={2021},
      eprint={2106.09685},
      archivePrefix={arXiv},
      primaryClass={cs.CL},
      url={https://arxiv.org/abs/2106.09685}, 
}

@misc{nie2025llada,
      title={Large Language Diffusion Models}, 
      author={Shen Nie and Fengqi Zhu and Zebin You and Xiaolu Zhang and Jingyang Ou and Jun Hu and Jun Zhou and Yankai Lin and Ji-Rong Wen and Chongxuan Li},
      year={2025},
      eprint={2502.09992},
      archivePrefix={arXiv},
      primaryClass={cs.CL},
      url={https://arxiv.org/abs/2502.09992}, 
}

@misc{grover2025mercury,
      title={Mercury: Ultra-Fast Language Models Based on Diffusion}, 
      author={Inception Labs and Samar Khanna and Siddhant Kharbanda and Shufan Li and Harshit Varma and Eric Wang and Sawyer Birnbaum and Ziyang Luo and Yanis Miraoui and Akash Palrecha and Stefano Ermon and Aditya Grover and Volodymyr Kuleshov},
      year={2025},
      eprint={2506.17298},
      archivePrefix={arXiv},
      primaryClass={cs.CL},
      url={https://arxiv.org/abs/2506.17298}, 
}

@misc{ye2025dream,
      title={Dream 7B: Diffusion Large Language Models}, 
      author={Jiacheng Ye and Zhihui Xie and Lin Zheng and Jiahui Gao and Zirui Wu and Xin Jiang and Zhenguo Li and Lingpeng Kong},
      year={2025},
      eprint={2508.15487},
      archivePrefix={arXiv},
      primaryClass={cs.CL},
      url={https://arxiv.org/abs/2508.15487}, 
}

@inproceedings{kanarya2b,
    title = "Mukayese: {T}urkish {NLP} Strikes Back",
    author = "Safaya, Ali  and
      Kurtulu{\c{s}}, Emirhan  and
      Goktogan, Arda  and
      Yuret, Deniz",
    editor = "Muresan, Smaranda  and
      Nakov, Preslav  and
      Villavicencio, Aline",
    booktitle = "Findings of the Association for Computational Linguistics: ACL 2022",
    month = may,
    year = "2022",
    address = "Dublin, Ireland",
    publisher = "Association for Computational Linguistics",
    url = "https://aclanthology.org/2022.findings-acl.69",
    doi = "10.18653/v1/2022.findings-acl.69",
    pages = "846--863",
}

@misc{kumru2b,
  title={Kumru},
  author={Turker, Meliksah and Ari, Erdi and Han, Aydin},
  year={2025},
  url={https://huggingface.co/vngrs-ai/Kumru-2B}
}

@misc{uludogan2024turna,
      title={TURNA: A Turkish Encoder-Decoder Language Model for Enhanced Understanding and Generation}, 
      author={Gökçe Uludoğan and Zeynep Yirmibeşoğlu Balal and Furkan Akkurt and Melikşah Türker and Onur Güngör and Susan Üsküdarlı},
      year={2024},
      eprint={2401.14373},
      archivePrefix={arXiv},
      primaryClass={cs.CL},
      url={https://arxiv.org/abs/2401.14373}, 
}

@misc{taori2023alpaca,
  author = {Rohan Taori and Ishaan Gulrajani and Tianyi Zhang and Yann Dubois and Xuechen Li and Carlos Guestrin and Percy Liang and Tatsunori B. Hashimoto },
  title = {Stanford Alpaca: An Instruction-following LLaMA model},
  year = {2023},
  publisher = {GitHub},
  journal = {GitHub repository},
  howpublished = {\url{https://github.com/tatsu-lab/stanford_alpaca}},
}

@misc{schweter2020berturk,
  author       = {Stefan Schweter},
  title        = {{BERTurk - BERT models for Turkish (1.0.0)}},
  year         = {2020},
  howpublished = {\url{https://doi.org/10.5281/zenodo.3770924}},
  note         = {Zenodo}
}

@misc{sahoo2024simpleeffectivemaskeddiffusion,
      title={Simple and Effective Masked Diffusion Language Models}, 
      author={Subham Sekhar Sahoo and Marianne Arriola and Yair Schiff and Aaron Gokaslan and Edgar Marroquin and Justin T Chiu and Alexander Rush and Volodymyr Kuleshov},
      year={2024},
      eprint={2406.07524},
      archivePrefix={arXiv},
      primaryClass={cs.CL},
      url={https://arxiv.org/abs/2406.07524}, 
}

@misc{li2025rethinking,
      title={Rethinking Multilingual Continual Pretraining: Data Mixing for Adapting LLMs Across Languages and Resources}, 
      author={Zihao Li and Shaoxiong Ji and Hengyu Luo and Jörg Tiedemann},
      year={2025},
      eprint={2504.04152},
      archivePrefix={arXiv},
      primaryClass={cs.CL},
      url={https://arxiv.org/abs/2504.04152}, 
}

@misc{marone2025mmbertmodernmultilingualencoder,
      title={mmBERT: A Modern Multilingual Encoder with Annealed Language Learning}, 
      author={Marc Marone and Orion Weller and William Fleshman and Eugene Yang and Dawn Lawrie and Benjamin Van Durme},
      year={2025},
      eprint={2509.06888},
      archivePrefix={arXiv},
      primaryClass={cs.CL},
      url={https://arxiv.org/abs/2509.06888}, 
}

@misc{toraman2024llamaturkadaptingopensourcegenerative,
      title={LlamaTurk: Adapting Open-Source Generative Large Language Models for Low-Resource Language}, 
      author={Cagri Toraman},
      year={2024},
      eprint={2405.07745},
      archivePrefix={arXiv},
      primaryClass={cs.CL},
      url={https://arxiv.org/abs/2405.07745}, 
}

@misc{instructurca,
  author={Duygu Altinok},
  title={InstrucTurca: A Diverse Instructional Content Dataset for Turkish},
  year={2024}
}

@misc{dllm,
    author = {Zhanhui Zhou and Lingjie Chen and Hanghang Tong and Dawn Song},
    title = {dLLM: Simple Diffusion Language Modeling},
    year = {2025},
    publisher = {GitHub},
    journal = {GitHub repository},
    howpublished = {\url{https://github.com/ZHZisZZ/dllm}},
}

@misc{touvron2023llamaopenefficientfoundation,
      title={LLaMA: Open and Efficient Foundation Language Models}, 
      author={Hugo Touvron and Thibaut Lavril and Gautier Izacard and Xavier Martinet and Marie-Anne Lachaux and Timothée Lacroix and Baptiste Rozière and Naman Goyal and Eric Hambro and Faisal Azhar and Aurelien Rodriguez and Armand Joulin and Edouard Grave and Guillaume Lample},
      year={2023},
      eprint={2302.13971},
      archivePrefix={arXiv},
      primaryClass={cs.CL},
      url={https://arxiv.org/abs/2302.13971}, 
}

@misc{er2025cetvelunifiedbenchmarkevaluating,
      title={Cetvel: A Unified Benchmark for Evaluating Language Understanding, Generation and Cultural Capacity of LLMs for Turkish}, 
      author={Yakup Abrek Er and Ilker Kesen and Gözde Gül Şahin and Aykut Erdem},
      year={2025},
      eprint={2508.16431},
      archivePrefix={arXiv},
      primaryClass={cs.CL},
      url={https://arxiv.org/abs/2508.16431}, 
}

@article{hardalov2020exams,
  title={EXAMS: Multi-subject High School Examinations for Cross-lingual QA},
  author={Hardalov, M. and Mihaylov, T. and Zlatkova, D. and Dinkov, Y. and Koychev, I. and Nvakov, P.},
  journal={arXiv preprint arXiv:2011.03080},
  year={2020}
}

@article{ozturk2021ironytr,
  title={IronyTR: Irony Detection in Turkish Informal Texts},
  author={Ozturk, Asli Umay and Cemek, Yesim and Karagoz, Pinar},
  journal={International Journal of Intelligent Information Technologies (IJIIT)},
  volume={17},
  number={4},
  pages={1--18},
  year={2021},
  publisher={IGI Global}
}

@inproceedings{amasyali2004news,
  title={Otomatik Haber Metinleri Sınıflandırma},
  author={Amasyalı, M. F. and Yıldırım, T.},
  booktitle={12th Signal Processing and Communications Applications Conference (SIU)},
  pages={224--226},
  year={2004},
  organization={IEEE}
}

@dataset{yilmaz_bilkent_turkish_writings_2025,
  title={Compilation of Bilkent Turkish Writings Dataset},
  author={Yilmaz, Selim F.},
  year={2025},
  publisher={Zenodo},
  doi={10.5281/zenodo.15498155},
  url={https://doi.org/10.5281/zenodo.15498155},
  version={2.0},
  note={Compilation of Turkish creative writings from Bilkent University Turkish 101 and 102 courses (2014-2025). 9,119 student writings collected and structured for NLP and educational research. Original content by Bilkent University students and instructors.}
}

@inproceedings{budur2020mnli,
  title={Data and Representation for Turkish Natural Language Inference},
  author={Budur, E. and Özçelik, R. and Güngör, T.},
  booktitle={Proceedings of EMNLP 2020},
  year={2020},
  publisher={Association for Computational Linguistics}
}

@misc{bandarkar2023belebele,
      title={The Belebele Benchmark: a Parallel Reading Comprehension Dataset in 122 Language Variants},
      author={Lucas Bandarkar and Davis Liang and Benjamin Muller and Mikel Artetxe and Satya Narayan Shukla and Donald Husa and Naman Goyal and Abhinandan Krishnan and Luke Zettlemoyer and Madian Khabsa},
      year={2023},
      eprint={2308.16884},
      archivePrefix={arXiv},
      primaryClass={cs.CL}
}

@inproceedings{fikri2021sts,
  title={Turkish Dataset for Semantic Textual Similarity},
  author={Fikri, F. B. and Oflazer, K. and Yanıkoğlu, B.},
  booktitle={29th Signal Processing and Communications Applications Conference (SIU)},
  pages={1--4},
  year={2021},
  doi={10.1109/SIU53274.2021.9477982}
}

@inproceedings{ponti2020xcopa,
  title={XCOPA: Multilingual Dataset for Causal Commonsense Reasoning},
  author={Ponti, E. M. and Glavaš, G. and Majewska, O. and Liu, Q. and Vulić, I. and Korhonen, A.},
  booktitle={Proceedings of EMNLP 2020},
  year={2020},
  url={https://ducdauge.github.io/files/xcopa.pdf}
}

@ONLINE{wikidump,
    author = "Wikimedia Foundation",
    title  = "Wikimedia Downloads",
    url    = "https://dumps.wikimedia.org"
}

@InProceedings{10.1007/978-3-031-70563-2_16,
author="Altinok, Duygu",
editor="N{\"o}th, Elmar
and Hor{\'a}k, Ale{\v{s}}
and Sojka, Petr",
title="Bella Turca: A Large-Scale Dataset of Diverse Text Sources for Turkish Language Modeling",
booktitle="Text, Speech, and Dialogue",
year="2024",
publisher="Springer Nature Switzerland",
address="Cham",
pages="196--213",
isbn="978-3-031-70563-2"
}

@inproceedings{Ortiz_Su_rez_2020,
   title={A Monolingual Approach to Contextualized Word Embeddings for Mid-Resource Languages},
   url={http://dx.doi.org/10.18653/v1/2020.acl-main.156},
   DOI={10.18653/v1/2020.acl-main.156},
   booktitle={Proceedings of the 58th Annual Meeting of the Association for Computational Linguistics},
   publisher={Association for Computational Linguistics},
   author={Ortiz Suárez, Pedro Javier and Romary, Laurent and Sagot, Benoît},
   year={2020} }

\appendix
\onecolumn
\section{Generation Examples}
\label{sec:appendix}

In this section, we present qualitative examples generated by \textit{Diffutron}. To demonstrate performance across varying lengths, we utilized two \texttt{LLaDAGeneratorConfig} settings as detailed in Table~\ref{tab:gen_params}.

\begin{table}[h]
\centering
\caption{Generation configurations. Common parameters apply to both settings.}
\label{tab:gen_params}
\begin{adjustbox}{width=\linewidth}
\begin{tabular}{l c c l}
\hline
\textbf{Config} & \textbf{Examples} & \textbf{Steps/Tokens} & \textbf{Common Parameters} \\
\hline
Longer Context & First 3 & 128 & \texttt{temp}=0.1, \texttt{block\_len}=32, \texttt{rep\_pen}=1.2, \\
Shorter Context & Last 4 & 64 & \texttt{remask}="low\_conf", \texttt{stochastic}=False, \texttt{cfg}=0.0 \\
\hline
\end{tabular}
\end{adjustbox}
\end{table}

\vspace{-0.3cm}

\begin{tcolorbox}[
    colback=white,
    colframe=black,
    arc=5pt,
    boxrule=0.5pt,
    left=8pt, right=8pt, top=6pt, bottom=6pt,
]

\textbf{Instruction:}  
Bugün hava nasıl?                                              

\vspace{6pt}

\textbf{Answer:}  
Yapay zeka dil modeli olarak gerçek zamanlı bilgilere erişimim yok. Ancak, en son hava durumu bilgilerini kontrol etmek için AccuWeather gibi bir haber kaynağı kullanabilir veya güncellemeler sağlayan yerel devlet web sitesine göz atabilirsiniz.

\end{tcolorbox}

\begin{tcolorbox}[
    colback=white,
    colframe=black,
    arc=5pt,
    boxrule=0.5pt,
    left=8pt, right=8pt, top=6pt, bottom=6pt,
]

\textbf{Instruction:}  
Bana kısa bir hikaye yaz.

\vspace{6pt}

\textbf{Answer:}  
Elbette, işte size basit ve kolay bir hikaye:

Bir zamanlar küçük bir kasabada yaşayan Lily adında genç bir kız varmış. Bir gün arkadaşları ve ailesiyle birlikte yürüyüşe çıkarken çok güzel bir manzarayla karşılaştı! Parkta rengarenk ağaçların olduğu büyük bir bahçeydi. Çok fazla renk vardı ama yine de yeşildi. Lily'nin gözleri muhteşem manzara karşısında şaşkına döndü ve ormanda oynamaya başladı. 
\end{tcolorbox}

\begin{tcolorbox}[
    colback=white,
    colframe=black,
    arc=5pt,
    boxrule=0.5pt,
    left=8pt, right=8pt, top=6pt, bottom=6pt,
]

\textbf{Instruction:}  
Kitap okumak insanlara ne kazandırır?

\vspace{6pt}

\textbf{Answer:}  
İnsanlar kitap okurken, farklı kültürler ve gelenekleri hakkında yeni şeyler öğrenebilirler. Ayrıca kişisel gelişimlerini geliştirebilecekleri için kendilerini daha iyi hissedebilirler. Bu bilgiyi öğrenmelerine yardımcı olmak için çevrimiçi olarak bulabilir veya başkalarıyla konuşabilirler. Ek olarak, kitap okuma bireylerin duygularını ve düşüncelerini ifade etmelerine olanak tanır; bu da onların fikirlerinin ve becerilerinin geliştirilmesine yol açar.
\end{tcolorbox}

\begin{tcolorbox}[
    colback=white,
    colframe=black,
    arc=5pt,
    boxrule=0.5pt,
    left=8pt, right=8pt, top=6pt, bottom=6pt,
]

\textbf{Instruction:}  
Fotosentez neden önemlidir?

\vspace{6pt}

\textbf{Answer:}  
Fotosentez, güneş ışığının karbondioksit ve suyu glikoza dönüştürdüğü süreçtir. Bu glikoz daha sonra enerji üretimi için bir besin kaynağı olarak kullanılır.
\end{tcolorbox}

\begin{tcolorbox}[
    colback=white,
    colframe=black,
    arc=5pt,
    boxrule=0.5pt,
    left=8pt, right=8pt, top=6pt, bottom=6pt,
]

\textbf{Instruction:}  
İnsanlar neden uyur?

\vspace{6pt}

\textbf{Answer:}  
Uyku, vücudun düzgün çalışması için gereklidir. Vücudumuzun iyi bir durumda kalmasını ve daha rahat hissetmesini sağlamaya yardımcı olur.
\end{tcolorbox}

\begin{tcolorbox}[
    colback=white,
    colframe=black,
    arc=5pt,
    boxrule=0.5pt,
    left=8pt, right=8pt, top=6pt, bottom=6pt,
]

\textbf{Instruction:}  
Film ve kitap arasındaki fark nedir?

\vspace{6pt}

\textbf{Answer:}  
Bir film genellikle bir kitabın hikayesini takip ederken, filmin konusu daha basittir. Kitaplar genellikle hikaye anlatımını içerirken filmler görsel uyarlamayı gerektirir.
\end{tcolorbox}

\begin{tcolorbox}[
    colback=white,
    colframe=black,
    arc=5pt,
    boxrule=0.5pt,
    left=8pt, right=8pt, top=6pt, bottom=6pt,
]

\textbf{Instruction:}  
İnsanlar neden spor yapar?

\vspace{6pt}

\textbf{Answer:}  
Spor, vücudun ısınmasına ve kasların güçlenmesine yardımcı olmak için egzersizdir. Ayrıca fiziksel aktiviteyi teşvik etmek ve genel sağlığı geliştirmek için de kullanılır.
\end{tcolorbox}

\end{document}